\documentclass[11pt,letterpaper]{article}
\usepackage{emnlp2016}
\usepackage{times}
\usepackage{latexsym}
\usepackage{natbib}
\usepackage{verbatim}
\usepackage{amsmath}
\usepackage{cases}
\usepackage{xcolor}
\usepackage{cleveref}
\usepackage{cancel}
\usepackage{url}

\emnlpfinalcopy

\def\emnlppaperid{***}

\newcommand{\aline}[1]{\textcolor{magenta}{#1}}

\newcommand{\marco}[1]{\textcolor{green}{#1}}

\renewcommand{\aline}[1]{#1}

\renewcommand{\marco}[1]{#1}

\title{Enhancing the LexVec Distributed Word Representation  Model \\
Using Positional Contexts and External Memory}

\author{Alexandre Salle\textsuperscript{1} \quad Marco Idiart\textsuperscript{2} \quad Aline Villavicencio\textsuperscript{1} \\
  \textsuperscript{1} Institute of Informatics \\
  \textsuperscript{2} Physics Department \\
  Universidade Federal do Rio Grande do Sul \\
  Porto Alegre, Brazil \\
  {\tt \{atsalle,avillavicencio\}@inf.ufrgs.br, idiart@if.ufrgs.br }}

\date{}

\begin{document}

\maketitle

\begin{abstract}
 \aline{In this paper we take a state-of-the-art model for distributed word representation that explicitly factorizes the positive pointwise mutual information (PPMI) matrix using window sampling and negative sampling and address two of its shortcomings. 
 We improve syntactic performance by using positional contexts, and solve the need to store the PPMI matrix in memory by working on aggregate data in external memory. The effectiveness of both modifications is shown using word similarity and analogy tasks. }
\end{abstract}

\section{Introduction}
Distributed word representations have become a mainstay in \emph{natural language processing},  %
enjoying a slew of applications \citep{Sebastiani2002,Turian2010,socher2013parsing}. 
Though \citet{Baroni2014} suggested that \emph{predictive} models which use neural networks to generate the distributed word representations (also known as \emph{embeddings} in this context) outperform \emph{counting} models which work on co-occurrence matrices, recent work shows evidence \aline{to the contrary} \citep{Levy2014linguisticregexplicit,salle2016}. 

In this paper, we focus on improving a state-of-the-art counting model, LexVec
(Salle et al., 2016), which performs
factorization of the \emph{positive pointwise mutual information} (PPMI) matrix using \emph{window sampling
and negative sampling} (WSNS). \citet{salle2016} suggest that LexVec matches and often outperforms competing models in  word similarity and semantic analogy tasks. Here we show that using positional contexts to approximate syntactic dependencies yields state-of-the-art performance on syntactic analogy tasks as well.
We also show how it is possible to approximate WSNS using aggregate data, eliminating random access to the PPMI matrix, enabling the use of external memory. Though not undertaken in this paper, this modification effectively allows LexVec to be trained on web-scale corpora.

This paper is organized as follows: we review the LexVec model ($\S$2) and detail how positional contexts and external memory can be incorporated into the model ($\S$3). We describe evaluation methods ($\S$4) and discuss results in terms of related work ($\S$5)  and finish with conclusions and future work.

Source code for the enhanced model is available at \url{https://github.com/alexandres/lexvec}.

\section{LexVec}
\label{sec:LexVec}
LexVec uses WSNS to factorize the PPMI matrix into two lower rank matrices.  The co-occurrence matrix $M$ is calculated from word-context pairs $(w, c)$ obtained
by sliding a symmetric window of size $win$ over the training corpus $C$.  The PPMI matrix is then calculated as follows

\vspace{-0.4cm}
\begin{footnotesize}
\begin{equation}
PPMI_{wc} = max(0, \log \frac{M_{wc} \; M_{**}}{ M_{w*} \; M_{*c} })
\end{equation}
\end{footnotesize}
\vspace{-0.4cm}

\noindent where $*$ represents index summation. 

The word and context embeddings $W$ and $\tilde{W}$, with dimensions $|V| \times d$ (where $V$ is the vocabulary and $d$ the embedding dimension), are obtained by the minimization via \emph{stochastic gradient descent} (SGD) of a combination of the loss functions

\vspace{-0.4cm}

\begin{footnotesize}
\begin{align}
\label{eq:lexvec2}
L_{wc} &= \frac{1}{2} (W_w\tilde{W_c}^\top - PPMI_{wc})^2 \\
\label{eq:lexvec3}
L_{w} &= \frac{1}{2} \sum\limits_{i=1}^k{\mathbf{E}_{w_i \sim P_n(w)} (W_w\tilde{W_{w_i}}^\top - PPMI_{ww_i})^2 }
\end{align}
\end{footnotesize}
\vspace{-0.4cm}

\noindent using WSNS. $P_n$ is the distribution used for drawing negative samples, chosen to be

\vspace{-0.4cm}
\begin{footnotesize}
\begin{align}
\label{eq:pneg}
P_n(w) =\#(w)^\alpha / \sum\limits_{w} \#(w)^\alpha
\end{align}
\end{footnotesize}

\vspace{-0.4cm}
\noindent with $\alpha = 3/4$ \citep{Mikolov2013negative,salle2016},
and $\#(w)$ the unigram frequency of $w$. 

\aline{Two methods were defined for the minimization of \cref{eq:lexvec2,eq:lexvec3}: \emph{Mini-batch} and \emph{Stochastic} \citep{salle2016}. Since the latter is more computationally efficient and yields equivalent results, we adopt it in this paper.} 
The Stochastic method extends the context window with noise words drawn using negative sampling according to \cref{eq:pneg}. The key idea is that window sampling is likely to sample related words, approximating their vectors using \cref{eq:lexvec2}, while negative sampling is likely to select unrelated words, scattering their vectors using \cref{eq:lexvec3}. The resulting global loss, where $\#(w,c) = M_{wc}$  is thus

\vspace{-0.4cm}
\begin{footnotesize}
\begin{align}
\begin{split}
\label{eq:lexvecstochasticglobal}
L &= \sum\limits_{(w,c)} \#(w,c) L_{wc} + \sum\limits_{w} \#(w) L_{w}
\end{split}
\end{align}
\end{footnotesize}
\vspace{-0.4cm}

Given a word $w$ and context word $c$, \cref{eq:lexvecstochasticglobal} is proportional to $\#(w)$ and $\#(c)$. This is the desired behaviour for the global loss function, since the more frequent $w$ or $c$ are in the corpus, the more confident we can be about the corpus estimated $PPMI_{wc}$. Suppose both $\#(w)$ and $\#(c)$ are high, but $PPMI_{wc}$ is low. This is unequivocal evidence of negative correlation between them, and so we should put more effort into approximating their $PPMI$. The argument is analogous for high $PPMI$.  If on the other hand $\#(w)$ and $\#(c)$ are low, we cannot be too confident about the  corpus estimated $PPMI_{wc}$, and so less effort should be spent on its approximation. 

\section{Enhancing LexVec}

\subsection{Positional Contexts}
As suggested \aline{by \cite{Levy2015improvingdist} and \cite{salle2016}}, positional contexts (introduced in \citet{Levy2014linguisticregexplicit}) are a potential solution to poor performance on syntactic analogy tasks. Rather than only accounting for which context words appear around a target word, positional contexts also account for their position relative to the target word. For example, in the sentence \emph{``the big \textbf{dog} barked loudly''}, target word $dog$ has contexts $(the^{-2}, big^{-1}, barked^1, loudly^2)$. The co-occurrence matrix, before having dimensions $|V| \times |V|$, takes on dimensions $|V| \times 2*win*|V|$ when using positional contexts.

This can be incorporated into LexVec with two minor modifications: 1) The context embedding $\tilde{W}$ \aline{takes} on dimensions $2*win*|V| \times d$, 2) Negative sampling must now sample positional contexts rather than simple contexts. This latter point requires that the distribution from which negative samples are drawn become

\vspace{-0.4cm}
\begin{footnotesize}
\begin{align}
\begin{split}
\label{eq:poscontextunigram}
P_n(c^i) = M_{*c^i}^\alpha /\sum\limits_{c^i} M_{*c^i}^\alpha
\end{split}
\end{align}
\end{footnotesize}
\vspace{-0.4cm}

Without positional contexts, either $W$ or $W + \tilde{W}$ can be used as embeddings. Since positional contexts make the dimensions of both matrices incompatible, $\tilde{W}$ cannot be used directly. We propose using the sum of all positional context vectors as the context vector for a word $( \tilde{W}^{pos} )$ .

\subsection{External Memory}
As window sampling scans over the training corpus and negative sampling selects random contexts, $(w,c^i)$ pairs are generated and the corresponding $PPMI_{wc^i}$ cell must be accessed so that 
\cref{eq:lexvec2,eq:lexvec3} can be minimized. Unfortunately, this results in random access to the $PPMI$ matrix which requires it \marco{to} be kept in main memory. \citet{Pennington2014} show that the under certain assumptions, this sparse matrix grows as $O(|C|^{0.8})$, which bounds the maximum corpus size that can be processed by LexVec.

\marco{We propose an approximation to WSNS that works as follows: All the word-context pairs $(w,c^i)$ generated by window sampling the corpus and by negative sampling each target word are first written to a file $F$. The file $F$ is then sorted with duplicated lines collapsed, and the lines written in the format $(w, c^i, +/-, tot, M_{wc^i})$, where $+/-$ indicates if the pair occurred or not in the corpus, $tot$ is the number of times the pair occurs including negative sampling, and $M_{wc^i}$ the number of times it occurred in the corpus.     }
$F$'s construction requires $O(|C|)$ external memory, and only $O(1)$ main memory. Additionally, all $M_{w*}$ and $M_{*c^i}$ are kept in main memory, using $O(|V|)$ space. This is nearly identical to the way in which GloVe builds its sparse co-occurrence matrix on disk, with the additional logic for adding and merging negatively sampled pairs.

We now present two ways to proceed with training: \emph{multiple iteration} or \emph{single iteration}.

\textbf{Multiple Iteration (MI)}: In this variant, $F$ is shuffled. For each tuple $(w, c^i, +/-, tot, M_{wc^i})$ in $F$, \cref{eq:lexvec2} is minimized $tot$ times, using $M_{w*}$, $M_{*c^i}$, and $M_{wc^i}$ to calculate $PPMI_{wc^i}$ if marker is $+$, else $PPMI_{wc^i}$ is equal to zero.   

\textbf{Single Iteration (SI)}: For every tuple $(w, c^i, +/-, tot, M_{wc^i})$ in $F$, write the tuple $(w, c^i, +/-, 1, M_{wc^i})$ $tot$ times to a new file $F'$. Then shuffle $F'$ and execute the MI algorithm described above on $F'$.

Both MI and SI are minimizing the exact same global loss function given by \cref{eq:lexvecstochasticglobal} as LexVec without external memory, the only difference between the three being the order in which \emph{word-context} pairs are processed.

\begin{table*}[t]
  \centering
  \begin{footnotesize}
  \begin{tabular}{c|ccccccccc}
\hline
Method & WSim & WRel & MEN & MTurk & RW & SimLex-999 & MC & RG & SCWS \\
\hline
PPMI-SVD	&	.731	& .617	& .731	& .627	& .427	& .303	& .770	& .756	& .615 \\
GloVe	&	.719 & .607	& .736	& .643	& .400	& .338	& .725	& .774	& .573 \\
SGNS	&	.770	& .670	& \textbf{.763}	& \textbf{.675}	& \textbf{.465}	& .339	& .823	& .793	& .643 \\
\hline
LexVec + Pos. + $W$	&	.740	& .631	& .744	& .645	& .464	&	\textbf{.358}	& .784	&	.775	& \textbf{.651} \\
LexVec + Pos. + $(W + \tilde{W}^{pos})$	&	.763	& \textbf{.676}	& .758	& .652	& .458	&	.333	& .781	&	.811	& .634 \\
LexVec + $W$	&	.741	& .622	& .733	& .628	& .457	&	.338	& .820	&	.808	& .638 \\
LexVec + $(W + \tilde{W})$	&	.763	& .671	& .760	& .655	& .458	& .336	& .816	& \textbf{.827}	& .630 \\
\hline
LexVec + Pos. + SI + $W$	&	.763	& .636	& .740	& .636	& .456	&	.356	& \textbf{.829}	&	.779	& .646 \\
LexVec + Pos. + SI  + $(W + \tilde{W}^{pos})$	&	\textbf{.776}	& \textbf{.676}	& .753	& .643	& .453	&	.333	& .811	&	.808	& .631 \\
LexVec + SI  + $W$	&	.741	& .634	& .730	& .619	& .444	&	.323	& .816	&	.780	& .621 \\
LexVec + SI + $(W + \tilde{W})$	&	.766	& .672	& .751	& .630	& .454	&	.322	& .818	&	.803	& .615 \\
\hline
LexVec + Pos. + MI + $W$	&	.745	& .636	& .727	& .639	& .414	&	.314	& .801	&	.787	& .635 \\
LexVec + Pos. + MI + $(W + \tilde{W}^{pos})$	&	.762	& .648	& .744	& .653	& .439	&	.325	& .804	&	\textbf{.827}	& .636 \\
LexVec + MI + $W$	&	.696	& .595	& .712	& .591	& .421	&	.322	& .814	&	.800	& .607 \\
LexVec + MI + $(W + \tilde{W})$	&	.739	& .646	& .750	& .639	& .448	&	.334	& .800	&	.790	& .629 \\
\hline
  \end{tabular}
  \end{footnotesize}
  \caption{Spearman rank correlation on word similarity tasks. Pos. = using positional contexts}
  \label{tab:wordsim}
\end{table*}

\begin{table*}[t]
  \centering
  \begin{footnotesize}
  \begin{tabular}{c|ccc}
\hline
Method & \begin{tabular}[x]{@{}c@{}}GSem\\3CosAdd / 3CosMul\end{tabular} & \begin{tabular}[x]{@{}c@{}}GSyn\\3CosAdd / 3CosMul\end{tabular} & \begin{tabular}[x]{@{}c@{}}MSR\\3CosAdd / 3CosMul\end{tabular} \\
\hline
PPMI-SVD	& .460 / .498 &  .445 / .455	&  .303 / .313 \\
GloVe &	\textbf{.818} / .813	& .630 / .626	& .539 / \textbf{.547} \\
SGNS	&	.773 / .777	& .642 / .644	& .481 / .505 \\
\hline
LexVec + Pos. + $W$ & .808 / .810	&	.633 / \textbf{.658} & .496 / .526 \\
LexVec + Pos. + $(W + \tilde{W}^{pos})$ & .799 / .808	&	.585 / .597 & .408 / .444 \\
LexVec + $W$ & .787 / .782	&	.597 / .613 & .445 / .475 \\
LexVec + $(W+\tilde{W})$ &	.794 / .807	&	.543 / .555 & .378 / .408 \\
\hline
LexVec + Pos. + SI + $W$	& .783 / .782	&	.611 / .630 & .456 / .484 \\
LexVec + Pos. + SI + $(W + \tilde{W}^{pos})$	& .801 / .810	&	.576 / .586 & .389 / .416 \\
LexVec + SI + $W$	& .760 / .766	&	.528 / .536 & .338 / .370 \\
LexVec + SI + $(W + \tilde{W})$	& .771 / .791	&	.450 / .473 & .268 / .300 \\
\hline
LexVec + Pos. + MI + $W$	& .624 / .620	&	.505 / .488 & .336 / .321 \\
LexVec + Pos. + MI + $(W + \tilde{W}^{pos})$	& .713 / .713	&	.561 / .555 & .385 / .384 \\
LexVec + MI + $W$	& .584 / .584	&	.384 / .369 & .187 / .180 \\
LexVec + MI + $(W + \tilde{W})$	& .697 / .707	&	.491 / .481 & .290 / .298 \\
\hline
  \end{tabular}
  \end{footnotesize}
  \caption{Results on word analogy tasks, given as percent accuracy.}
  \label{tab:analogies}
\end{table*}

\section{Materials}
\label{sec:materials}
We report results from \citet{salle2016} and use the same training corpus and parameters to train LexVec with positional contexts and external memory. The corpus is a Wikipedia dump from June 2015, tokenized, lowercased, and split into sentences, removing punctuation and converting numbers to words, for a final vocabulary of 302,203 words. 

All generated embeddings have dimensionality equal to 300. As recommended in \citet{Levy2015improvingdist} and used in \citet{salle2016}, the PPMI matrix used in all LexVec models and in PPMI-SVD is transformed using context distribution smoothing exponentiating context frequencies to the power $0.75$. PPMI-SVD is the singular value decomposition of the PPMI matrix. LexVec and PPMI-SVD use symmetric windows of size $2$. Both GloVe \citep{Pennington2014} and Skip-gram with negative sampling (SGNS) \citep{Mikolov2013negative} were trained using a symmetric window of size $10$. GloVe was run for $50$ iterations, using parameters $x_{max} = 100$, $\beta = 3/4$, and learning rate of $0.05$. LexVec and SGNS were run for 5 iterations, using 5 negative samples, and initial learning rate of $0.025$. LexVec, PPMI-SVD, and SGNS use dirty subsampling \citep{Mikolov2013negative,Levy2015improvingdist} with threshold $t = 10^{-5}$. Words in the training corpus with unigram probability $f$ greater than $t$ are discarded with probability $1 - \sqrt{t/f}$. For LexVec, we report results for $W$ and $(W + \tilde{W}^{pos})$ embeddings when using positional contexts, otherwise $W$ and $(W + \tilde{W})$. For PPMI-SVD and GloVe we report $(W + \tilde{W})$, and for SGNS, $W$, that correspond to their best results.

The goal of our evaluation is to determine whether: 1) Positional contexts improve syntactic performance 2) The use of external memory is a good approximation of WSNS. Therefore, we perform the exact same evaluation as \citet{salle2016}, namely the WS-353 
Similarity (WSim) and Relatedness (WRel) 
\citep{Finkelstein2001},  MEN \citep{bruni2012distributional}, MTurk \citep{radinsky2011word}, RW \citep{Luong2013}, SimLex-999 \citep{hill2015simlex}, MC \citep{Miller1991}, RG \citep{Rubenstein1965}, and SCWS \citep{Huang2012} word similarity tasks\footnote{http://www.cs.cmu.edu/~mfaruqui/suite.html}, and the Google semantic (GSem) and syntactic (GSyn) analogy \citep{Mikolov2013sgoriginalnonegative} and MSR syntactic analogy dataset \citep{Mikolov2013linguisticregcontinuous} tasks. Word similarity tasks use cosine similarity. Word analogy tasks are solved using both 3CosAdd and 3CosMul \citep{Levy2014linguisticregexplicit}.

\section{Results}
\label{sec-results}
Positional contexts improved performance in both similarity (\cref{tab:wordsim}) and analogy tasks (\cref{tab:analogies}). As hypothesized, their use significantly improved LexVec's performance on syntactic analogies, leading to the highest score on GSyn, surpassing GloVe and SGNS. This confirms the relevance of using positional contexts to capture syntactic dependencies.

\citet{salle2016} reported that combining word and context embeddings scored marginally higher in word similarity tasks, and that holds true in our experiments, even for $\tilde{W}^{pos}$.
In the analogy tasks, using only the word embedding leads to far better syntactic performance, indicating that the embedding $W$ strikes a better balance between syntax and semantics than does $W + \tilde{W}$.

The SI external memory implementation very closely approximates the \emph{standard} variant (without the use of external memory), which was expected given that they minimize the exact same loss function. The gap between MI and standard was much wider. It seems that there is value in the way WSNS uses corpus ordering of word-context pairs to train the model. The SI variant more closely mimics this order, distributing same pair occurrences over the entire training. MI, on the other hand, has a completely artificial ordering, distant from corpus and SI's ordering.

\section{Conclusion and Future Work}
This paper presented two improvements  \aline{to the word embedding model LexVec. The first yields state-of-the-art performance on syntactic analogies through the use of positional contexts. The second  solves the need to store the PPMI matrix in main memory by using external memory.   
The SI variant of the external memory  implementation was a good approximation of standard LexVec's WSNS, enabling future training using web-scale corpora.} 

In future work, %
we plan to explore the model's hyperparameter space, which could potentially boost model performance, having so far restricted ourselves to parameters recommended in \citet{Levy2015improvingdist}.
Finally, following 
\citet{tsvetkov2015evaluation}, we will pursue evaluation of the model on downstream tasks in addition to the intrinsic evaluations used in this paper.

\bibliography{serveboy_manual2,confs}
\begin{footnotesize}
\bibliographystyle{acl_natbib}
\end{footnotesize}
\appendix

\end{document}